\documentclass[authoryear,preprint,11pt]{elsarticle}

\usepackage{amssymb}
\usepackage{amsthm}
\usepackage{eurosym}
\usepackage{color}
\usepackage{bm}
\usepackage{booktabs}
\usepackage{multirow}
\usepackage{longtable}
\usepackage{arydshln}
\usepackage{geometry}
\usepackage{subcaption}
\usepackage{comment}
\usepackage{setspace}
\usepackage{acronym}

\usepackage[]{algorithm2e}
\usepackage{listings}
\usepackage{mathtools}
\usepackage{enumitem}
\newlist{steps}{enumerate}{1}
\setlist[steps, 1]{label = Step \arabic*}

\acrodef{MCDA}{Multiple Criteria Decision Analysis}
\acrodef{ROR}{Robust Ordinal Regression}
\acrodef{SMAA}{Stochastic Multicriteria Acceptability Analysis}
\acrodef{ERA}{Extreme Ranking Analysis}
\acrodef{DM}{Decision Maker}
\acrodef{MAVT}{Multi-Attribute Value Theory}
\acrodef{MILP}{Mixed-Integer Linear Programming}
\acrodef{LP}{Linear Programming}
\acrodef{MCMC}{Markov Chain Monte Carlo}
\acrodef{DRSA}{Dominance-based Rough Set Approach}

\journal{Information Sciences}
\usepackage{etoolbox}
\makeatletter
\patchcmd{\ps@pprintTitle}{\footnotesize\itshape
Preprint submitted to \ifx\@journal\@empty Elsevier
\else\@journal\fi\hfill\today}{\relax}{}{}
\makeatother

\usepackage{lineno, blindtext}

\begin{document}
\onehalfspacing

\begin{frontmatter}
    \title{Using a Segmenting Description approach in  \\ Multiple  Criteria Decision Aiding}
    \author[put]{Mi\l{}osz Kadzi\'{n}ski}
    \author[put]{Jan Badura}
    \author[ceg-ist]{Jos\'e Rui Figueira\corref{cor1}}\ead{figueira@tecnico.ulisboa.pt}
    \address[put]{Institute of Computing Science, Faculty of Computing, Pozna\'n University of Technology, Pozna\'n, Poland}
    \address[ceg-ist]{CEG-IST, Instituto Superior T\'{e}cnico,  Universidade de Lisboa, Portugal}
\cortext[cor1]{Corresponding author: CEG-IST, Instituto Superior T\'{e}cnico,  Universidade de Lisboa, Av. Rovisco Pais, 1, 1049-001 Lisboa, Portugal. Phone: +351 21 841 798 1. Fax: +351 21 841 797 9 (Date: March, 5th 2019).}

\begin{abstract}
\noindent
    We propose a new method for analyzing a set of parameters in a multiple criteria ranking method. Unlike the existing techniques, we do not use any optimization technique, instead incorporating and extending a Segmenting Description approach. While considering a value-based preference disaggregation method, we demonstrate the usefulness of the introduced algorithm in a multi-purpose decision analysis exploiting a system of inequalities that models the Decision Maker's preferences. Specifically, we discuss how it can be applied for verifying the consistency between the revealed and estimated preferences as well as for identifying the sources of potential incoherence. Moreover, we employ the method for conducting robustness analysis, i.e., discovering a set of all compatible parameter values and verifying the stability of suggested recommendation in view of multiplicity of feasible solutions. In addition, we make clear its suitability for generating arguments about the validity of outcomes and the role of particular criteria. We discuss the favourable characteristics of the Segmenting Description approach which enhance its suitability for use in Multiple Criteria Decision Aiding. These include keeping in memory an entire process of transforming a system of inequalities and avoiding the need for processing the inequalities contained in the basic system which is subsequently enriched with some hypothesis to be verified. The applicability of the proposed method is exemplified on a numerical study.
\end{abstract}
\vspace{0.25cm}
\end{frontmatter}



\section{Introduction}\label{sec:introduction}
\noindent Each \ac{MCDA} method requires the  specification of some technical or preference-based parameters~\citep{GEF16}. These parameters are used to derive recommendations concerning a~particular problem in view of multiple pertinent factors. A~direct specification of a~set of parameter values is often considered troublesome since it imposes a~high cognitive effort on the part of a~\ac{DM}~\citep{greco2008ordinal, Kadzinski13}. Such a difficulty may be alleviated through the use of preference disaggregation procedures~\citep{MG18}, which infer the parameter values using an~optimization technique based on the DM's holistic judgments concerning a small subset of reference alternatives~\citep{LagrezeSi01}. Even if such an~indirect specification of preference information is considered to be more user-friendly and less cognitively demanding, the comprehensive preferences estimated by the method may differ from the ones expressed by the DM. This can be due to different types of inconsistencies.

Dealing with inconsistent decision making is an important topic in behavioural operations research. For Daniel Kahneman -- the Noble prize winner in Economic Sciences -- and his collaborators: ``\emph{the problem is that humans are unreliable DMs; their judgments are strongly influenced by irrelevant factors, such as their current mood, the time since their last meal, and the weather}''~\citep{KahnemanEtAl16}. For this reason, inconsistency in the DMs' behaviour needs to be treated as an inherent part of the decision processes. Detecting why a~judgment is in some sense inconsistent allows for a co-constructive interaction between analysts and DMs to render clear, understandable decision aiding process for making better informed and justifiable decisions~\citep{Roy10}.

In MCDA~\citep{Roy96}, the phenomenon of inconsistent judgments appears even more frequently given the nature of multiple conflicting criteria. In fact, apart from the internal incoherence of the DM's judgment policy, the inconsistencies in \ac{MCDA} may arise due to the preferences provided by the DM is not aligned with the assumptions of a~selected preference model, the requirements of a~particular decision aiding method, and the context of a~specific application.
Yet the number of works for dealing with inconsistent judgments in MCDA is rather scarce. In particular, the general MCDA tools for identifying inconsistent judgments \citep[see, e.g.,][]{MousseauEtAl03, MousseauEtAl06, greco2008ordinal} are very much related to those used for \emph{infeasibility analysis} in single objective optimization~\citep[see, e.g.,][]{Chinneck08}. These tools were mainly designed to restore feasibility in \ac{LP} in case a~set of constraints leads to an empty hyper-polyhedron. The two most well-known problems considered in this context are the following:
    \begin{itemize}[label={--}]
        \item Find an \emph{irreducible infeasible set} (IIS) \citep[see, e.g.,][]{Chinneck94,vanLoon81} being a minimal set of constraints that lead to infeasibility.
        \item Find a \emph{maximum feasibility subset} (MFS) \citep[see, e.g.,][]{AmaldiEtAl99,Sankaran93} being a maximal feasible subset of constraints that do not contradict each other.
    \end{itemize}
The most common problem considered in MCDA in the presence of a~contradictory set of constraints~\citep{MousseauEtAl03} is to find an adequate number of (or, if possible, all) subsets of constraints underlying the infeasibility, usually prioritizing the subsets with the least cardinality. In general, such \emph{explanatory subsets} are not necessarily IISs nor MFSs. Instead, their role is to reveal and illustrate some sources of contraction, helping the DMs to better understand their judgments and to construct their preferences in a close interaction with the analysts. 

Another stream of research for dealing with inconsistency in preference disaggregation methods incorporates the optimization techniques to minimize a~difference between the expressed and estimated preferences relative to the values of preference model parameters~\citep{Beuthe01}. In particular, one can minimize either the maximal error~\citep{Despotis90} or the sum of errors~\citep{sisk2rice85} introduced in the mathematical constraints reconstructing the DM's holistic judgments by means of the preference model to be inferred. If there remain some ``errors'' in the estimation, some reference judgments of the DM cannot be reproduced by the model. Overall, the model is accepted if the highlighted differences are not substantial in the eyes of DM. Nevertheless, this way of proceeding can imply some validity problems~\citep{ROY1987297}.

More recently, the optimization techniques for dealing with inconsistency were incorporated for other purposes, which are not linked directly to minimizing the difference between the DM's preference information and the results provided by the method. These purposes include verification of some hypotheses regarding the validity of results~\citep{ROY2010629} and generation of justifications of such outcomes~\citep{Kadzinski14ORSP}. Specifically, in the context of preference disaggregation methods, some \ac{LP} models were solved within the scope of robustness analysis, verifying the stability of results that can be obtained in view of multiplicity of compatible values of preference model parameters~(see, e.g., \cite{greco2008ordinal, Figueira09, DIAS2002332, Kadzinski15, KADZINSKI2016167}). Yet other algorithms were introduced to show a~direct link between the robust results and subsets of DM's preference information pieces implying their truth or falsity~\citep{Kadzinski14ORSP}.

In this paper, we propose a~method that allows to analyze inconsistent judgments, provide robust conclusions, and generate explanations/argumentation without an optimization model. This is a~strong missing link in the MCDA literature. The introduced approach is memory-based, helping to identify all steps made by the DM leading to the inconsistent judgments or justifying the (in)validity of a~verified hypothesis. Such capabilities are of great help within a~co-constructive interactive preference learning approach.

The proposed method incorporates and extends a~Segmenting Description (SD) approach originally proposed by~\cite{Roy63, Roy70}. The SD algorithm aims to rewrite an original system of inequalities, with a fixed number of decision variables, into an equivalent ``SD system''.  This is attained by means of three fundamental operations, called isolating, crossover, and adjunction. In this new system, the variables are indexed and each inequality has the form of either a lower- or an upper-bound for the variable with the highest index appearing in it. The newly created system has the same set of feasible solutions as the original one, and any solution can be obtained by a~successive choice of the values for variables with the increasing indices. The SD method is theoretically different from the existing methods based on linear or mixed integer (with $0-1$ auxiliary variables) \ac{LP}. Its advantage derives from the possibility of finding all sources of inconsistency by backtracking, since it keeps in memory the whole process.

The SD algorithm fits perfectly within the disaggregation paradigm~\cite[see, e.g.,][]{LagrezeSi82,LagrezeSi01}, which is represented in our paper by the family of UTA methods~\citep{Siskos05}. They are used to support the solution of multiple criteria ranking problems by estimating an additive value function based on the DM's pairwise comparisons of reference alternatives. An intermediate step in the estimation process consists in formulating a~system of linear inequalities brought out by some normalization and monotonicity constraints and DM's indirect preferences~\citep{LagrezeSi82}. The variables contained in the model are bounded, and decide upon the shape of marginal value functions and/or trade-off weights assigned to the particular criteria. Such a system of inequalities defines a~set of additive value functions compatible with the DM's preference information~\citep{greco2008ordinal}. We show how the SD method can be applied to such a~system to conduct a~multi-purpose analysis, exploiting the DM's pairwise comparisons and a respective set of additive value functions.

Specifically, we consider the problem of detecting the reasons underlying the inconsistency between the DM's holistic judgments and an \emph{a~priori} assumed preference model without an optimization technique~\citep{MousseauEtAl03, KADZINSKI2018472}. Then, we show how to verify the truth of the necessary and possible preference relations, being confirmed by, respectively, all or at least one value functions compatible with the DM's judgments (assuming that her/his preference information is consistent with the model)~\citep{greco2008ordinal, Figueira09}.
In addition, we indicate how the SD approach can be used for computing the preference reducts and constructs. These are defined as, respectively, a minimal subset of DM's pairwise comparisons justifying the truth of some currently observable result and a maximal subset of such preference information pieces for which the currently non-observable result would be validated~\citep{Kadzinski14ORSP}. Finally, we introduce the concept of a \emph{holistic preference criteria reduct} denoting a minimal subset of criteria that -- when incorporated into an~assumed preference model -- would reproduce the DM's holistic judgments. The latter idea is inspired by the \ac{DRSA}~\citep{DRSA01}, where the notion of a reduct was traditionally considered in view of violating the dominance relation by the DM's preferences, though without incorporating any model of knowledge~\citep{SUSMAGA201445}. In this perspective, we also reveal the usefulness of the SD approach in understanding why some subsets of criteria are not capable of reproducing the DM's preferences.

The paper is organized in the following way. In Section~\ref{sec:section_application}, we define a system of inequalities aligning with the assumption of UTA. Section~\ref{sec:section_ds_approach} provides a theoretical background and discuses the properties of the Segmenting Description approach. In Section~\ref{sec:section_experiments_results}, we demonstrate the use of Segmenting Description for dealing with inconsistency, robustness analysis, and generating explanations on a~numerical example. Section~\ref{sec:conslusions_remarks} concludes and provides avenues for future research.

\section{Definition of the system of inequalities for the UTA method}\label{sec:section_application}
\noindent In this section, we define a system of inequalities based on the ranking procedure incorporated in the UTA method~\citep{LagrezeSi82}. Let $A = \{a_1, \ldots, a_i,\ldots, a_m\}$ denote a set of alternatives evaluated in terms of $n$ criteria $ G = \{g_1,\ldots, g_j,\ldots,g_n\}$, where $g_j: A \rightarrow \mathbb{R}$. We assume, without loss of generality, that all criteria are of gain-type, i.e., the greater $g_j(a_i)$, the better alternative $a_i$ is on criterion $g_j$, for $j = 1,\ldots, n$. The performances on $g_j$ are bounded within the interval $\lbrack g_{j,*}, g_{j}^* \rbrack$.

\vfill\newpage

\noindent \textbf{Preference model.} UTA incorporates an additive value function as a preference model, hence aggregating the performances of each alternative $a_i \in A$ on multiple criteria into a single score~\citep{Keeney76, KADZINSKI2017146}:
 \begin{equation}
    \label{math:additive_function}
    U(a_i) = \sum_{j=1}^{n} u_{j}(a_i) \in[0,1],
 \end{equation}
where $u_j$, $j=1,\ldots,n$, are non-decreasing marginal value functions defined through a set of $\gamma_j$~characteristic points $g_j^s$, $s=1,\ldots,\gamma_j$, equally distributed in the domain of performances $\lbrack g_{j,*}, g_{j}^* \rbrack$.

\vspace{0.5cm}

\noindent \textbf{Preference information.} The preference model is estimated by UTA based on the DM's holistic judgments concerning a subset of reference alternatives $A^R \subseteq A$. The preference ($\succ_{DM}$) and indifference ($\sim_{DM}$) relations imposed by the DM's indirect preference information are translated into parameters of a compatible value function. The system of linear inequalities related with the compatibility with the DM's pairwise comparisons contains the following constraints:
 \begin{flalign}
    & u_j(g_{j,*}) = 0, \quad j = 1,\ldots,n, \label{utacon1} \\
    & \sum_{j=1}^{n} u_j(g_{j}^*) = 1, \label{utacon2}  \\
    & u_j(g_j^s) \geqslant u_j(g_j^{s-1}),\;\text{for}\;j=1,\ldots,n\;\text{and}\;s=2,\ldots,\gamma_j, \label{utacon3} \\
    & u_j(a_i)=u_j(g_j^{s-1})+\frac{(u_j(g_j^s)-u_j(g_j^{s-1}))(g_j(a_i)-g_j^{s-1})}{(g_j^s-g_j^{s-1})}, \nonumber\\
    &\;\;\;\;\;\;\;\;\;\;\;\;\;\;\;\; \text{for}\;g_j(a_i) \in [g_j^{s-1},g_j^s), \; j=1,\ldots,n, \mbox{ and } a\in A^R, \label{utacon4} \\
    & \sum_{j=1}^{n} u_{j}(a_i) > \sum_{j=1}^{n} u_{j}(a_k), \mbox{ for } a_i, a_k \in A^R, \; a_i \succ_{DM} a_k, \label{utacon5} \\
    & \sum_{j=1}^{n} u_{j}(a_i) = \sum_{j=1}^{n} u_{j}(a_k), \mbox{ for }  a_i, a_k \in A^R, \; a_i \sim_{DM} a_k. \label{utacon6}
 \end{flalign}
Constraints~(\ref{utacon1}) and (\ref{utacon2}) normalize an additive value function within $\lbrack 0, 1\rbrack$. Constraints~(\ref{utacon3}) and~(\ref{utacon4}) ensure that marginal value functions are non-decreasing and piece-wise linear. Constraints~(\ref{utacon5}) and~(\ref{utacon6}) reproduce the pairwise comparisons provided by the DM.

In case marginal value functions are assumed to be linear (i.e., when $\gamma_j = 2$, $j=1,\ldots,n$), the system of inequalities cab be defined as follows:
 \begin{flalign}
    & w_j \geqslant 0, \quad j = 1,\ldots,n, \label{2utacon1} \\
    & \sum_{j=1}^{n} w_j = 1, \label{2utacon2}  \\
    & \sum_{j=1}^{n} w_{j} \frac{g_j(a_i)-g_j^1}{g_j^{\gamma_j}-g_j^1} > \sum_{j=1}^{n} w_{j} \frac{g_j(a_k)-g_j^1}{g_j^{\gamma_j}-g_j^1}, \mbox{ for } a_i, a_k \in A^R, \; a_i \succ_{DM} a_k, \label{2utacon5} \\
    & \sum_{j=1}^{n} w_{j} \frac{g_j(a_i)-g_j^1}{g_j^{\gamma_j}-g_j^1} = \sum_{j=1}^{n} w_{j} \frac{g_j(a_k)-g_j^1}{g_j^{\gamma_j}-g_j^1}, \mbox{ for }  a_i, a_k \in A^R, \; a_i \sim_{DM} a_k, \label{2utacon6}
 \end{flalign}
where $w_j = u_j(g_{j}^*)$, $j=1,\ldots,n$, corresponding to a maximal share of criterion $g_j$ in the comprehensive value can be interpreted as a trade-off weight associated with $g_j$.

On the one hand, if the system of inequalities was contradictory, the DM's pairwise comparisons cannot be reproduced by the UTA model. This means that there is no additive value function that would be able to rank the reference alternatives as desired by the DM. In this case, the DM should be informed about the reasons underlying an~inconsistency so that to treat it in an~adequate way, e.g., by revising his or her holistic judgments on a~subset of reference alternatives. On the other hand, if the system of inequalities was non-contradictory, the set of compatible additive value models is non-empty. This set should be exploited in a way that provides outcomes and explanations contributing to the recommendation regarding the ranking problem.

\section{The Segmenting Description approach}
\label{sec:section_ds_approach}
\label{sec:subsection_princ_and_objec}
\noindent In this section, we present the SD approach~\citep{Roy63, Roy70}, which is incorporated within the proposed method for a multi-purpose decision analysis. The algorithm processes an original system of $r$~inequalities involving $s$~variables, and aims to rewrite into a slightly redundant ``SD system''. It considers the variables which are bounded from below and above, being indexed from $x_1$ to $x_s$. Each inequality has the form of either a lower- or an upper-bound for the variable with the highest index appearing in it.

After the transformation performed by the SD approach, the variable with the lowest index in the SD system, $x_1$, is bounded by constants, and all remaining variables $x_k$, $k \in \{2,\ldots, s\}$, are bounded by functions $f_k(x_{1},\ldots,x_{k-1})$ involving the variables with indices not greater than $k-1$. In addition, a set of feasible solutions remains unchanged. Any solution can be obtained by a~successive choice of the values for variables with increasing indices $x_1, x_2, \ldots, x_s$.

\subsection{Fundamental operations}\label{sec:subsection_fund_steps}
\noindent The SD approach is made operational via three fundamental operations:

\begin{itemize}[label={--}]
    \item Isolating (ISO) operation which modifies a given inequality by isolating the variable with the highest index $k$ and creating a lower- or an upper-bound for variable $x_{k}$ in the following way: $l_{k}(x_{1},\ldots, x_{k-1}) \leqslant x_{k}$ or $x_{k} \leqslant u_{k}(x_{1},\ldots, x_{k-1})$.
    \item Crossover (CRO) operation which combines a lower- and an upper-bound concerning the same variable $x_{k}$ into a~new inequality $l_{k}(x_{1},\ldots, x_{k-1}) \leqslant u_{k}(x_{1},\ldots, x_{k-1})$.
    \item Adjunction (ADJ) operation which enriches the SD system with the inequality obtained by applying the ISO operation to the inequality created by the CRO operation. Note that adding such an inequality does not change the set of feasible solutions. The ADJ operation may involve additional steps that ensure that the new inequality is not redundant in comparison with the ones already contained in the system (e.g., inequality $x_{1} \geqslant 0.3$ is redundant in view of inequality $x_{1} \geqslant 0.5$ already contained in the system). This topic is further discussed in Section~\ref{sec:redundant_inequalities}.
\end{itemize}

\subsection{The main steps of the SD method}\label{sec:algorithm_steps}
\noindent The Segmenting Description method consists of the following steps:
\begin{enumerate}
    \item Process the initial system by transforming each equality into the following pair of inequalities:
    \[ x = y \implies \left\{\def\arraystretch{1.2}%
        \begin{array}{@{}l@{\quad}l@{\quad}l@{}}
            x \geqslant y \\
            x \leqslant y \\
        \end{array}\right.
    \] and each strict inequality into a~non-strict constraint involving an~arbitrarily small slack variable~$\epsilon$:
    \[ x > y \implies x \geqslant y + \epsilon, \]
    \[ x < y \implies x + \epsilon \leqslant y. \]
    \item Index the variables in the following way: $x_1, x_2,\ldots, x_s$. Although the SD algorithm works correctly with any indexing, it is recommended to index the variables so that the more often variable is present in the initial system of inequalities, the lower its index should be. Such an ordering should reduce the number of CRO operations that the algorithm has to perform.
    \item Apply the ISO operation to each inequality of the initial system.
    \item Process variable $x_s$ as follows: \label{step:process_variable}
    \begin{enumerate}
        \item[(\textit{a})]  Identify a lower-bound $l_{k}(x_{1},\ldots, x_{k-1})$ of $x_s$.
        \item[(\textit{b})]  Identify an upper-bound $u_{k}(x_{1},\ldots, x_{k-1})$ of $x_s$.
        \item[(\textit{c})]  Perform the CRO operation on the identified inequalities.
        \item[(\textit{d})]  If the new inequality contradicts the existing ones, terminate by presenting the results of the method.
        \item[(\textit{e})]  Apply the ADJ operation on the new inequality.
        \item[(\textit{f})] Repeat sub-steps (\textit{a})-(\textit{e}) for each combination of the lower and upper-bounds.
    \end{enumerate}
    \item Process variables $x_{s-1}, x_{s-2}, \ldots, x_{1}$ using the operations contained in Step~\ref{step:process_variable}. \label{step:process_remaining_variables}
    \item Terminate by presenting the SD system.
\end{enumerate}

\noindent Note that it is possible to add new inequalities to the existing SD system without restarting the process from the beginning. Specifically, step \ref{step:process_remaining_variables} can be performed starting from variable $x_k$ with $k$ being the highest index present in the new inequalities. The CRO operation should be performed only for the combinations involving lower- or upper-bounds which are new in the system.

\subsection{Identifying contradictions}
\label{sec:identifying_contradiction}
\noindent The SD method keeps in memory the history of creating an SD system by labeling the inequalities. On the one hand, special labels are generated for the inequalities contained in the original system. On the other hand, the new inequalities obtained with the CRO operation are not only labelled, but one also stores their parents, i.e., the respective lower- and upper-bounds. Overall, the SD method employs the following labelling procedure:

\begin{itemize}[label={--}]
    \item All $r$ inequalities from the original system are indexed from $1$ to $r$. In addition, the parents of each inequality $\ell = 1, \ldots, r$ are indexed as $\ell, \ell$. Hence, the employed label has the following form: $\{\ell,\ell,\ell\}$.
    \item Each inequality created by the CRO operation is indexed with $\ell$, being the last index used by the method incremented by one. Furthermore, its parents are indexed with $h$ and $k$, being, respectively, the lower- and upper-bound used in the CRO operation. As a result, the label has the following form: $\{\ell,h,k\}$.
    \item It is not necessary to index the inequalities identified as redundant by the ADJ operation. This may, however, lead to neglecting some subset of contradictory inequalities.
\end{itemize}

\noindent When a contradictory inequality appears as a result of performing the CRO operation, it is possible to recursively follow its parents until reaching inequalities from the original system (labeled as $\{\ell,\ell,\ell\}$) to indicate which inequalities implied the contradiction. Such a~procedure generates a~single set of contradictory inequalities. To find all such sets, it is necessary to apply the procedure so that it processes all contradictions.

\subsection{Algorithmic issues and implementations details} \label{sec:section_impl_details}
\noindent This subsection presents the SD algorithm, its theoretical complexity, and how to deal with redundant inequalities.

\subsubsection{Main loop}\label{sec:main_loop}
\noindent A general idea underlying the SD algorithm, implementing the method presented in Section~\ref{sec:section_ds_approach}, can be represented as Algorithm~\ref{alg:croloop}. Let us denote by
$S$ a three dimensional array that stores inequalities. In particular, $S[\ell][LB][\cdot]$ contains all lower-bounds and $S[\ell][UB][\cdot]$ all upper-bounds for variable $x_\ell$. Each inequality is stored as a vector of coefficients. Hence, the CRO operation consists of the following subtraction: $l_{k}(x_{1},\ldots, x_{k-1}) - u_{k}(x_{1},\ldots, x_{k-1}) \leqslant 0$. The inequality is considered contradictory when coefficients associated with all variables are equal to zero and constant $b$ is positive, leading to $b \leqslant 0$.

\vspace{0.5cm}

\begin{algorithm}[H]
 \caption{Processing the CRO operations.}
 \label{alg:croloop}
\For{$\ell = s, \ldots, 1$}{
 \For{$L \in S[\ell][LB][\cdot]$}{
    \For{$U \in S[\ell][UB][\cdot]$}{
        {\tt constraint} $\leftarrow$ $CRO(L, U)$\;
         {\tt constraint} $\leftarrow$ $ISO({\tt constraint})$\;
        \If{{\tt constraint} is not contradictory}{
            $ADJ(S, {\tt tmp})$\;
        }{
            {\bf return}(contradictions)\;
        }
    }
 }
}
\end{algorithm}

\subsubsection{Complexity analysis}\label{sec:complexity}
\noindent The complexity of the ISO and CRO operations is linear with respect to the number of variables in the system. Those, however, are the only steps whose complexity is easy to be measured.  The three nested loops presented in Algorithm~\ref{alg:croloop} are expected to run in the exponential time with respect to the number of variables $s$ and the number of inequalities $r$ in the initial system. The values of coefficients also contribute to the complexity. When processing lower- and upper-bound of variable~$x_k$, the newly obtained inequality will be lower- or upper-bound of any variable $x_1,\ldots, x_{k-1}$. On the one hand, if~the index of new inequality is low, e.g., we obtain a bound for $x_1$, this inequality will be processed fewer times and create less additional inequalities.  On the other hand, if the index is high (e.g., $x_{k-1}$), the inequality will indirectly create many new inequalities. The computation time also relies heavily on how the inequalities spread between lower- and upper-bounds. Specifically, the more uniformly inequalities spread between these two types of bounds, the more CRO operations will be performed. If we assumed that all of $r$ initial inequalities are bounds for variable $x_s$ spread evenly between lower- and upper-bounds and after performing each CRO operation for some $x_k$ we achieved bound for variable $x_{k-1}$, and those also evenly spread between the lower- and upper-bounds, the complexity could be approximated by
$O(2^{2-2^s} r^{2^{s-1}})$.

\subsubsection{Redundant inequalities}\label{sec:redundant_inequalities}
\noindent Another important point is the ADJ operation. The simplest approach is to add all new inequalities to the system. This is, however, not always necessary. When not including the redundant inequalities in the system, the running time of the algorithm can be greatly reduced. Let us note that if this is done, we cannot identify all sets of contradictory inequalities.

Some inequalities can be removed from the system of inequalities without changing the feasible region. Since the SD algorithm creates new inequalities from the ones already existing in the system, each inequality created by the CRO operation is, by definition, redundant in comparison with the entire system. This, however, is not the case if we only considered a~subsystem that was influenced by the new inequality. If a new inequality is a bound for variable $x_k$, we can check whether it is redundant in the subsystem containing bounds for variables $x_1,\ldots, x_k$.

In the literature, we can find many methods for identifying redundancy in linear systems. In particular, \cite{S2010} discussed the following techniques: bounds method~\citep{brearley1975analysis}, linear programming method~\citep{caron1989degenerate}, deterministic approach~\citep{telgen1983identifying}, and heuristic method~\citep{paulraj2006heuristic}.

\section{Illustrative study: the use of Segmenting Description for dealing with inconsistency, robustness analysis, and generating explanations}
\label{sec:section_experiments_results}
\noindent In this section, we illustrate how the SD algorithm can contribute to a~better organization of a~decision aiding process in the context of multiple criteria problems. Specifically, we will apply the SD algorithm to a~numerical example, and use it for dealing with inconsistency in the provided preference information, conducting robustness analysis, and generating explanations of the provided recommendation.

Let us consider a problem originally formulated by~\cite{greco2008ordinal}. It concerns selection of a~new international sales manager by a~medium size firm producing some agricultural tools. The company considers $15$ potential candidates who are treated as decision alternatives: $a_1$--$a_{15}$. They were evaluated in terms of the following three criteria on a~$\lbrack 0, 100 \rbrack$ scale: sales management experience~($g_1$), international experience~($g_2$), and human qualities~($g_3$). The performance matrix is presented in Table~\ref{tab:uta_example}a.

\begin{table}[!ht]
\centering\footnotesize
\caption{Performance table used in the problem of sales manager selection (a) and constraint set induced by the preference information provided by the DM in the first iteration (b).}
\label{tab:uta_example}
\begin{tabular}{c c c c c r l}
 \hline
 \multicolumn{4}{c}{(a) Performances} &&  \multicolumn{2}{l}{(b) Constraints induced by the preference model} \\
 $a_i$ & $g_1$ & $g_2$ & $g_3$ & & \multicolumn{2}{l}{and DM's preference information} \\
 \hline
    $a_1$ & 4 & 16 & 63 & &$\lbrack C1\rbrack$ & $w_1 \geqslant 0$ \\
    $a_2$ & 28 & 18 & 28 &&  $\lbrack C2\rbrack$ &  $w_2 \geqslant 0$\\
    $a_3$ & 26 & 40 & 44 &&  $\lbrack C3\rbrack$ &  $w_3 \geqslant 0$\\
    $a_4$ & 2 & 2 & 68 &&  $\lbrack C4\rbrack$ &  $w_1 + w_2 + w_3 \geqslant 1$  \\
    $a_5$ &18 & 17 & 14 &&  $\lbrack C5\rbrack$ &  $w_1 + w_2 + w_3 \leqslant 1$  \\
    $a_6$ & 35 & 62 & 25 &&  $\lbrack C6\rbrack$ &  $0.56w_1 + w_2 + 0.25 w_3 \geqslant 0.15 w_1 + w_2 + 0.88 w_3$\\
    $a_7$ & 7 & 55 & 12  &&  $\lbrack C7\rbrack$ &  $0.56w_1 + w_2 + 0.25 w_3 \leqslant 0.15 w_1 + w_2 + 0.88 w_3$ \\
    $a_8$ & 25 & 30 & 12  &&  $\lbrack C8\rbrack$ &  $0.15 w_1 + w_2 + 0.88 w_3 \geqslant 0.40 w_1 + 0.47 w_2 + 0.12 w_3 + 0.01$\\
    $a_9$ & 9 & 62 & 88 &&  $\lbrack C9\rbrack$ &  $0.40 w_1 + 0.47 w_2 + 0.12 w_3 \geqslant w_1 + 0.68 w_2 + 0.01$\\
    $a_{10}$ & 0 & 24 & 73 &&  $\lbrack C10\rbrack$ &  $w_1 + 0.68 w_2 \geqslant 0.11 w_1 + 0.88 w_2 + 0.12 w_3 + 0.01$\\
    $a_{11}$ & 6 & 15 & 100 \\
    $a_{12}$ & 16 & 9 & 0 \\
    $a_{13}$ & 26 & 17 & 17 \\
    $a_{14}$ & 62 & 43 & 0 \\
    $a_{15}$ & 1 & 32 & 64 \\ \hline
\end{tabular}
\end{table}

\vfill\newpage

\subsection{Dealing with inconsistency in preference information}\label{sec:infeasible_example}
\noindent Let us assume that the DM (Chief Executive Officer of the company) attended a~few interviews with the candidates. This allowed him to provide the following reference ranking involving some holistic judgments concerning a~subset of decision alternatives:
\begin{equation}
    \label{refinfo:lv_ii}
    a_6 \sim a_9 \succ a_8 \succ a_{14} \succ a_7.
\end{equation}
When modelling the DM's preference information, we will assume that the marginal value functions are linear and $\epsilon = 0.01$. For this reason, the only variables contained in the set of constraints induced by the reference ranking are criteria weights: $w_1$, $w_2$, and $w_3$. The system of inequalities that will be exploited with the SD algorithm is presented in Table~\ref{tab:uta_example}b. Constraints $\lbrack C1\rbrack$--$\lbrack C5\rbrack$ are implied by the normalization of comprehensive values within a~$\lbrack 0,1\rbrack$ range, whereas constraints $\lbrack C6\rbrack$--$\lbrack C10\rbrack$ refer to the pairwise comparisons between the consecutive alternatives contained in the reference ranking.

We aim at verifying whether the DM's preference information is consistent with an~assumed preference model. Let us first apply the ISO operation to the constraint set from Table~\ref{tab:uta_example}b. The operation consists in isolating the variable with the highest index so that to create a~bound expressed as a~function of variables with lower indices and/or constants. The resulting inequalities labelled from $\{1,1,1\}$ to $\{10,10,10\}$ are presented in Table~\ref{example_iso}. This set involves eight lower- or upper-bounds for $w_3$, and just a~single lower-bound for $w_1$ and $w_2$. In addition, the inequalities derived from the reference ranking are associated with the respective pairwise comparison in the table. For example, $\{6,6,6\}$ and $\{7,7,7\}$ are implied by an~indifference between $a_6$ and $a_9$, whereas $\{10,10,10\}$ originates from $a_{14} \succ a_7$.

\begin{table}[!ht]
\centering\footnotesize
\caption{Initial system of labelled inequalities modelling the DM's preference information in the first iteration.}
\label{example_iso}
\begin{tabular}{lc | lcc}
 \hline
 Inequality & Label & Inequality & Label & Pairwise comparison \\
 \hline
$w_1 \geqslant 0$ & \{1,1,1\} & $w_3 \leqslant 0.67 w_1$ & \{6,6,6\} & $a_6 \sim a_9$  \\
$w_2 \geqslant 0$ & \{2,2,2\} & $w_3 \geqslant 0.67 w_1$ & \{7,7,7\} & $a_6 \sim a_9$\\
$w_3 \geqslant 0$ & \{3,3,3\} & $w_3 \geqslant -0.70 w_2 + 0.34 w_1 + 0.01$ & \{8,8,8\} & $a_9 \succ a_8$ \\
$w_3 \geqslant -w_2 - w_1 + 1$ & \{4,4,4\} & $w_3 \geqslant 1.80 w_2 + 4.97 w_1 + 0.08$ & \{9,9,9\} & $a_8 \succ a_{14}$\\
$w_3 \leqslant -w_2 - w_1 + 1$ & \{5,5,5\} & $w_3 \leqslant -1.67 w_2 + 7.39 w_1 - 0.08$ & \{10,10,10\} & $a_{14} \succ a_7$\\ \hline
\end{tabular}
\end{table}

 The next step consists in conducting the CRO operations for the inequalities involving each variable, starting with the one with the highest index. When it comes to variable $w_3$, it is delimited by $5$~lower- and $3$~upper-bounds (see Table~\ref{example_iso}). Thus, the crossover operation leads to $15$ new inequalities (see Table~\ref{example_crow3}).  All but one concern $w_2$, whereas $\{12,3,6\}$ that combined $w_3 \geqslant 0$ ($\{3,3,3\}$) and $w_3 \leqslant 0.67 w_1$ ($\{6,6,6\}$) constrains variable $w_1$.

\begin{table}[!ht]
\centering\footnotesize
\caption{The inequalities derived from an application of the CRO operation to variable $w_3$ in the first iteration.}
\label{example_crow3}
\begin{tabular}{ c c c c }
 \hline
 Inequality & Label & Inequality & Label \\ \hline
$w_2 \leqslant - w_1 + 1.0$ & $\{11,3,5\}$ & $w_2 \leqslant - 4.49 w_1 + 3.31$ & $\{18,8,5\}$  \\
$w_1 \geqslant 0$ & $\{12,3,6\}$ &  $w_2 \geqslant - 0.46 w_1 + 0.02$ & $\{19,8,6\}$  \\
$w_2 \leqslant  4.44 w_1 - 0.05$ & $\{13,3,10\}$ & $w_2 \leqslant  7.31 w_1 - 0.10$ & $\{20,8,10\}$   \\
$w_2 \geqslant - 1.67 w_1 + 1.0$ & $\{14,4,6\}$ & $w_2 \leqslant - 2.13 w_1 + 0.33$ & $\{21,9,5\}$ \\
$w_2 \leqslant  12.59 w_1 - 1.63$ & $\{15,4,10\}$  & $w_2 \leqslant - 2.39 w_1 - 0.05$ & $\{22,9,6\}$\\
$w_2 \leqslant - 1.67 w_1 + 1$ & $\{16,7,5\}$  &  $w_2 \leqslant  0.70 w_1 - 0.05$ & $\{23,9,10\}$\\
$w_2 \leqslant  4.04 w_1 - 0.05$ & $\{17,7,10\}$ \\
\hline
\end{tabular}
\end{table}

 As far as variable $w_2$ is concerned, it is delimited by $3$~lower- and $10$~upper-bounds. The CRO operation applied to these inequalities resulted in $29$ bounds for variable $w_1$ (see Table~\ref{example_crow2}). Note that a~crossover involving $\{14,4,6\}$ and $\{16,7,5\}$ did not produce any new constraint. For clarity of presentation, we include in Table~\ref{example_crow2} all constraints including the ones which are redundant.

\begin{table}[!ht]
\centering\footnotesize
\caption{The inequalities derived from an application of the CRO operation to variable $w_2$ in the first iteration.}
\label{example_crow2}
\begin{tabular}{ c c c c c c}
 \hline Inequality & Label & Inequality & Label & Inequality & Label \\ \hline
$w_1 \leqslant 1$ & \{24,2,11\} & $w_1 \geqslant 0$ & \{34,14,11\}&  $w_1 \geqslant 0.014$ & \{44,19,13\}\\
$w_1 \geqslant 0.011$ & \{25,2,13\}& $w_1 \geqslant 0.172$ & \{35,14,13\}&  $w_1 \geqslant 0.126$ & \{45,19,15\}\\
$w_1 \geqslant 0.129$ & \{26,2,15\}& $w_1 \geqslant 0.184$ & \{36,14,15\}&  $w_1 \leqslant  0.817$ & \{46,19,16\}\\
$w_1 \leqslant  0.600$ & \{27,2,16\}& $w_1 \geqslant 0.184$ & \{37,14,17\}&  $w_1 \geqslant 0.015$ & \{47,19,17\}\\
$w_1 \geqslant 0.012$ & \{28,2,17\}& $w_1 \leqslant  0.81$7 & \{38,14,18\}& $w_1 \leqslant  0.817$ & \{48,19,18\}\\
$w_1 \leqslant  0.737$ & \{29,2,18\}& $w_1 \geqslant 0.123$ & \{39,14,20\} &  $w_1 \geqslant 0.015$ & \{49,19,20\}\\
$w_1 \geqslant 0.013$ & \{30,2,20\}&  $w_1 \leqslant  - 1.452$ & \{40,14,21\} &  $w_1 \leqslant  0.185$ & \{50,19,21\}\\
$w_1 \leqslant  0.153$ & \{31,2,21\}& $w_1 \leqslant  - 1.453$ & \{41,14,22\} &  $w_1 \leqslant  - 0.033$ &\{51,19,22\}\\
$w_1 \leqslant  - 0.019$ & \{32,2,22\}& $w_1 \geqslant 0.444$ & \{42,14,23\} & $w_1 \geqslant 0.057$ &\{52,19,23\}\\
$w_1 \geqslant 0.069$ & \{33,2,23\}& $w_1 \leqslant  1.833$ & \{43,19,11\} \\
\hline
\end{tabular}
\end{table}

\begin{figure}[!h]
    \centering
    \includegraphics[scale=0.5]{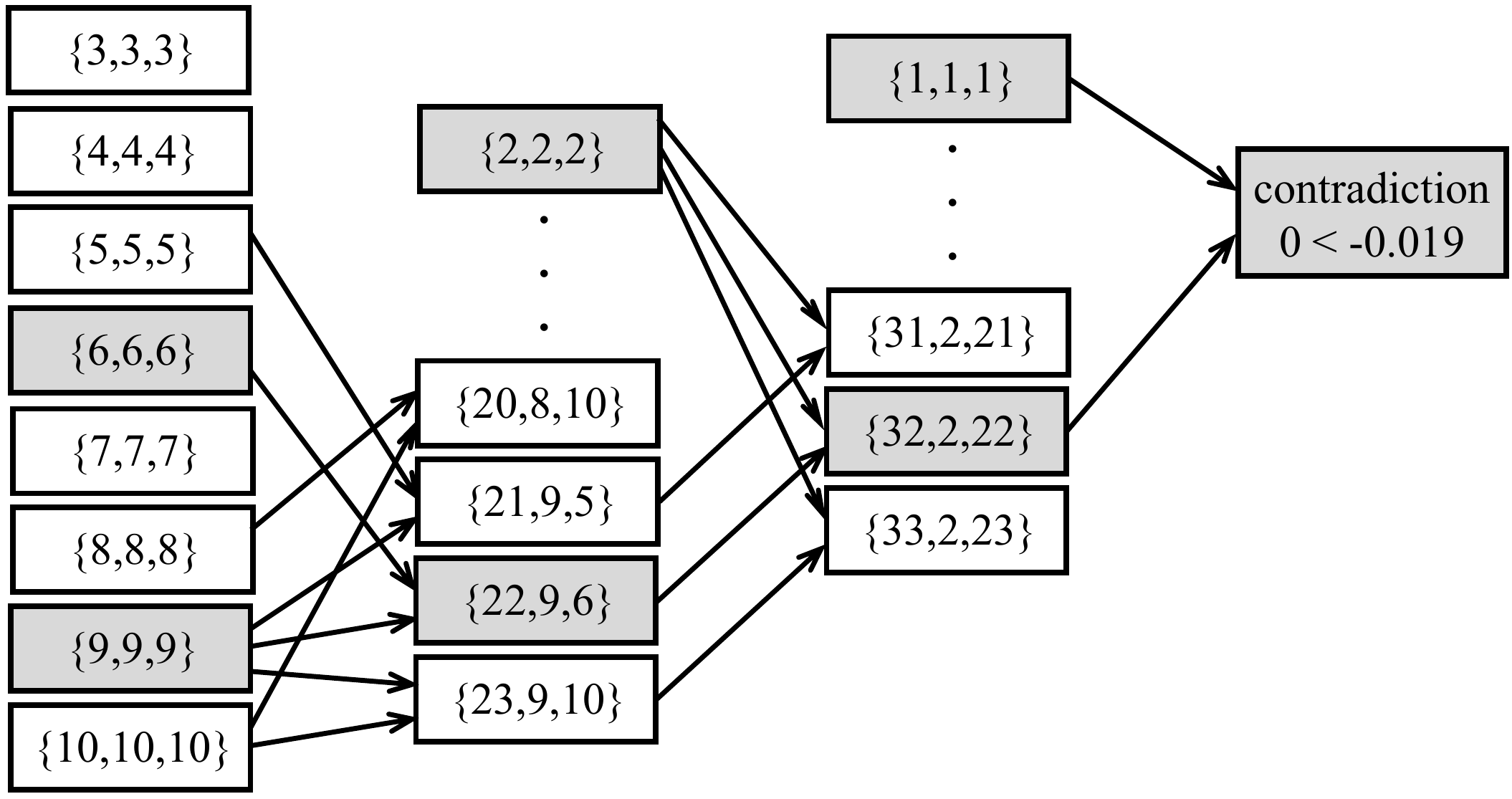}
    \caption{An example process of identifying the reasons underlying contradiction with the labelling procedure.}
    \label{fig:uta_linear_contradiction}
\end{figure}

Inequalities contained in Table~\ref{example_crow2} involve several contradictions. Their analysis by means of the backtracking procedure leads to identification of contradictory constraints in the initial system. The latter ones are, in turn, linked to the DM's preference information. In Figure~\ref{fig:uta_linear_contradiction}, we present such a~process for $w_1 \geqslant 0$ and $w_1 \leqslant  - 0.019$. It indicates that pairwise comparisons $a_6 \sim a_9$ ($\{6,6,6\}$) and $a_8 \succ a_{14}$ ($\{9,9,9\}$) cannot be considered jointly with the assumption on non-negativity of $w_1$ ($\{1,1,1\}$) and $w_2$ ($\{2,2,2\}$). Hence, the DM needs to revise his or her reference ranking so that it could be reproduced with an additive value function where per-criterion attractiveness is modelled with marginal linear functions.

\subsection{Identification of compatible preference model instances}
\label{sec:uta_example_feasible}
\noindent The use of the SD algorithm for dealing with inconsistency in preference information led to identification of the troublesome pairwise comparisons that cannot be reconstructed together with an assumed preference model. Let us assume that being presented with the two pairwise comparisons underlying incompatibility, the DM decided to remove $a_{14}$ from his or her reference ranking, hence eliminating the following pairwise comparison: $a_8 \succ a_{14}$. As a~result, the ranking involves four reference alternatives compared as follows:
\begin{equation}
    \label{refinfo:linear_1}
    a_6 \sim a_9 \succ a_8 \succ a_7.
\end{equation}
This reference ranking is translated into the respective set of constraints analogously as in Section~\ref{sec:infeasible_example}. Having applied the ISO operation to such a set and labelled the inequalities, we obtain a~system presented in Table~\ref{example_init_uta_linear}.

\begin{table}[!ht]
\caption{Initial system of labelled inequalities modelling the DM's preference information in the second iteration.}
\label{example_init_uta_linear}
\centering\footnotesize

\begin{tabular}{lc | lc}
 \hline
 Inequality & Label & Inequality & Label \\
 \hline
$w_1 \geqslant 0$ & \{1,1,1\} & $w_3 \leqslant 0.67 w_1$ & \{6,6,6\} \\
$w_2 \geqslant 0$ & \{2,2,2\} & $w_3 \geqslant 0.67 w_1$ & \{7,7,7\} \\
$w_3 \geqslant 0$ & \{3,3,3\} & $w_3 \geqslant - 0.70 w_2 + 0.34 w_1 + 0.01$ & \{8,8,8\} \\
$w_3 \geqslant - w_2 - w_1 + 1$ & \{4,4,4\} & $w_2 \leqslant 0.70 w_1 - 0.02$ & \{9,9,9\} \\
$w_3 \leqslant - w_2 - w_1 + 1$ & \{5,5,5\} \\ \hline
\end{tabular}
\end{table}


The CRO operations involving lower- and upper-bounds of variable $w_3$ introduced only 5 new inequalities (see Table~\ref{example_crow3_it2}). This is because the CRO operations combining \{4,4,4\} with \{5,5,5\} or \{6,6,6\} with \{7,7,7\} led to the redundant inequalities $0 \leqslant 0$, whereas crossover involving \{3,3,3\} and \{6,6,6\} implied $0 \leqslant w_1$ that is already present in the system.

\begin{table}[!ht]
\caption{The inequalities derived from an application of the CRO operation to variable $w_3$.}
\centering\footnotesize

\label{example_crow3_it2}
\begin{tabular}{ c c c c }
 \hline
 Inequality & Label & Inequality & Label \\ \hline
   $w_2 \leqslant - w_1 + 1$ & \{10,3,5\} &   $w_2 \leqslant - 4.49 w_1 + 3.31$ & \{13,8,5\}  \\
   $w_2 \geqslant - 1.67 w_1 + 1.0$ & \{11,4,6\} & $w_2 \geqslant - 0.46 w_1 + 0.02$ & \{14,8,6\}  \\
   $w_2 \leqslant - 1.67 w_1 + 1$ & \{12,7,5\} \\
\hline
\end{tabular}
\end{table}

 The CRO operations for variable $w_2$ created several lower- and upper-bounds for variable $w_1$ in form of constants. To save space, we do not list them all. Let us just note that they did not involve any contradiction. In turn, we present the final SD system (see Table~\ref{example_init_uta_linear_final}) with the constant bounds being either the highest lower-bounds or the least upper-bounds. Overall, the SD algorithm successfully processed the system of constraints derived from the DM's reference ranking, the alternatives' performances, and the assumed preference model. It means that the space of value functions compatible with the DM's indirect judgments is non-empty.

\begin{table}[!ht]
\caption{The SD system corresponding to the preference information provided in the second iteration.}
\centering\footnotesize

\label{example_init_uta_linear_final}
\begin{tabular}{rclc}
 \hline
 Inequality & Label & Inequality & Label \\
 \hline
$0 \leqslant w_3$ & \{3,3,3\}& $w_3 \leqslant - w_2 - w_1 + 1$ & \{5,5,5\} \\
$ - w_2 - w_1 + 1  \leqslant w_3$ & \{4,4,4\} & $w_3 \leqslant 0.67 w_1$ & \{6,6,6\} \\
$0.67 w_1  \leqslant w_3$ & \{7,7,7\}  \\
$ - 0.70 w_2 + 0.34 w_1 + 0.01  \leqslant w_3$ & \{8,8,8\} \\
\hline
$0 \leqslant w_2$ & \{2,2,2\} &  $w_2 \leqslant 0.70 w_1 - 0.02$ & \{9,9,9\}\\
$- 1.67 w_1 + 1.0  \leqslant w_2$ & \{11,4,6\} & $w_2 \leqslant - w_1 + 1$ & \{10,3,5\}\\
$-0.46 w_1 + 0.02  \leqslant w_2$ & \{14,8,6\} &  $w_2 \leqslant - 1.67 w_1 + 1$ & \{12,7,5\} \\
&&$w_2 \leqslant - 4.49 w_1 + 3.31$ & \{13,8,5\} \\
\hline
$0 \leqslant w_1$ & \{1,1,1\} & $w_1 \leqslant 0.60$ & \{15,2,12\} \\
$0.43 \leqslant w_1$ & \{16,11,9\} & & \\
\hline

\end{tabular}
\end{table}

 The analysis of the SD system supports understanding the interrelations between different feasible model parameters. In particular,  Table~\ref{example_init_uta_linear_final} indicates that $w_1$ attains values between $0.43$ and $0.6$. The bounds for the remaining variables can be expressed in function of the variables with lower indices. For example, for weight $w_2$ these bounds can be formulated as follows:
\begin{flalign*}
w_2 \geqslant \max \{0; - 1.67 w_1 + 1.0; -0.46 w_1 + 0.02 \};\\
w_2 \leqslant \min \{0.70 w_1 - 0.02; - w_1 + 1; - 1.67 w_1 + 1; - 4.49 w_1 + 3.31\}.
\end{flalign*}

\vfill\newpage

Thus, selection of a value for $w_1$ within the admissible range influences the values of other parameters. On the one hand, when assuming the lowest allowed value of $w_1$, the weights assigned to the remaining criteria are very much alike ($w_2 = 0.28$ and $w_3 = 0.29$). On the other hand, when $w_1$ takes the greatest admissible value, $g_3$ also has a significant impact on the comprehensive value ($w_3 = 0.4$) while $g_2$ becomes negligible ($w_2 = 0$).

The SD algorithm may be further used to analyze the DM's preference model more in-depth. Specifically, we can learn lower- and upper-bounds of each variable by conducting the SD algorithm with different indexing of variables, implying various processing orders. The computations that need to be performed are fairly similar to those presented in this section. Hence we will not list all inequalities, and solely provide the bounds for all variables that were derived from such an extended analysis:
\begin{flalign*}
0.43 \leqslant w_1 \leqslant 0.6, \; 0 \leqslant w_2 \leqslant 0.28, \mbox{ and } 0.29 \leqslant w_3 \leqslant 0.4.
\end{flalign*}

These bounds reveal that criterion $g_1$ has the greatest share in the comprehensive value. In fact, the least possible value of $w_1$ is greater than maximal admitted weights for $g_2$ and $g_3$.
Such a space of compatible weights not only makes evident all feasible solutions, but it can be exploited either by selecting a single representative set of weights or accounting for all such weights simultaneously. The former would lead to a~complete ranking of alternatives, whereas the latter would allow to verify the stability of the recommendation in view of indeterminacy of the DM's preference model due to the incompleteness of provided preference information.

\subsection{Robustness analysis}\label{sec:robustness}
\noindent In this section, we demonstrate how the SD algorithm can be used to conduct the robustness analysis~\citep{ROY2010629}. It consists in taking into account an uncertainty related to the multiplicity of compatible preference model instances~\citep{Kadzinski13, KADZINSKI2014211, KADZINSKI2016321}. In particular, we verify the stability of results that can be obtained for different pairs of alternatives for all such instances by verifying the truth of necessary and possible relations. Although we will present the outcomes for all pairs, we will discuss them in detail just for a~selected pair. When using the SD algorithm, we will use a~basis in form of an~initial system of labelled inequalities presented in Table~\ref{example_init_uta_linear}. This system will be extended with various hypotheses allowing to check the validity of robust preference relations.

\subsubsection{Necessary preference relation}
\label{sec:necessary_weak}
\noindent
The necessary preference relation holds for a given pair of alternatives if one of them is at least as good as the other for all compatible preference model instances~\citep{greco2008ordinal}. The first step of using the SD algorithm for verifying whether relation $a_i \succsim^N a_k$ is true consists of adding an inequality $U(a_k) > U(a_i)$ to the initial non-contradictory system modelling the DM's preferences. If the final system would involve some contradiction, then $U(a_k) > U(a_i)$ does not hold for any compatible preference model instance, and hence $a_i \succsim^N a_k$. Otherwise, $a_k$ is strictly preferred to $a_i$ for at least one compatible preference model instance, and thus $not(a_i \succsim^N a_k)$.

Let us first verify the truth of $a_{1} \succsim^N a_{14}$. For this purpose, we need to extend an initial system from Table~\ref{example_init_uta_linear} with constraint $U(a_{1}) < U(a_{14})$ corresponding to the following inequality $w_3 \leqslant 0.71 w_2 + 1.48 w_1 - 0.02$ (\{10,10,10\}).
The system obtained after the ISO and CRO operations is presented in Table~\ref{uta_necessary_false} (to save space we include only the extreme bounds for variable $w_1$). Its feasibility means that $U(a_{1}) < U(a_{14})$ can be instantiated for at least one compatible preference model instance, and hence $a_{1} \succsim^N a_{14}$ does not hold.

\begin{table}[!ht]
\centering\footnotesize
\caption{The SD system for verifying the truth of necessary preference relation for pair $a_{1}$ and $a_{14}$ ($a_{1} \succsim^N a_{14}$).}
\label{uta_necessary_false}
\begin{tabular}{lclc}
 \hline
 Inequality & Label & Inequality & Label \\
 \hline
    $w_3 \geqslant 0$ & \{3,3,3\} & $w_3 \leqslant - w_2 - w_1 + 1$ & \{5,5,5\} \\
    $w_3 \geqslant - w_2 - w_1 + 1$ & \{4,4,4\} & $w_3 \leqslant + 0.67 w_1$ & \{6,6,6\} \\
    $w_3 \geqslant 0.67 w_1$ & \{7,7,7\} & $w_3 \leqslant + 0.71 w_2 + 1.48 w_1 - 0.02$ & \{10,10,10\} \\
    $w_3 \geqslant - 0.7 w_2 + 0.34 w_1 + 0.01$ & \{8,8,8\} & & \\
    \hline
    $w_2 \geqslant 0$ & \{2,2,2\} & $w_2 \leqslant + 0.7 w_1 - 0.02$ & \{9,9,9\} \\
    $w_2 \geqslant - 2.08 w_1 + 0.02$ & \{13,3,10\} & $w_2 \leqslant - w_1 + 1$ & \{11,3,5\} \\
    $w_2 \geqslant - 1.67 w_1 + 1.0$ & \{14,4,6\} & $w_2 \leqslant - 1.67 w_1 + 1$ & \{16,7,5\} \\
    $w_2 \geqslant - 1.45 w_1 + 0.59$ & \{15,4,10\} & $w_2 \leqslant - 4.49 w_1 + 3.31$ & \{18,8,5\} \\
    $w_2 \geqslant - 1.15 w_1 + 0.02$ & \{17,7,10\} & & \\
    $w_2 \geqslant - 0.46 w_1 + 0.02$ & \{19,8,6\} & & \\
    $w_2 \geqslant - 0.81 w_1 + 0.02$ & \{20,8,10\} & & \\
    \hline
    $w_1 \geqslant 0$ & \{1,1,1\} & $w_1 \leqslant + 0.60$ & \{23,2,16\}\\
    $w_1 \geqslant + 0.43$ & \{29,14,9\} & & \\ \hline
\end{tabular}
\end{table}

As far as the truth of $a_{14} \succsim^N a_{1}$ is concerned, its verification starts with extending an initial system (see Table~\ref{example_init_uta_linear}) with constraint $U(a_{14}) < U(a_{1})$ corresponding to the following inequality $w_3 \geqslant  0.71 w_2 + 1.48 w_1 + 0.02$ (\{10,10,10\}). The final system derived by the SD algorithm is presented in Table~\ref{uta_necessary_true} (we exhibit all inequalities, because this system will be subsequently exploited to compute the preferential reducts).
Note that \{2,2,2\} and \{18,10,6\} created an upper-bound for variable $w_1 \leqslant -0.02$ (\{24,2,18\}), which contradicts one of the model constraints $w_1 \geqslant 0$ (\{1,1,1\}). This means that $U(a_{1}) < U(a_{14})$ does not hold for any compatible preference model instance, which, in turn, implies $a_{14} \succsim^N a_{1}$.

\begin{table}[!ht]
\centering\footnotesize
\caption{The SD system for verifying the truth of necessary preference relation for pair $a_{14}$ and $a_{1}$ ($a_{14} \succsim^N a_{1}$).}
\label{uta_necessary_true}
\begin{tabular}{lclc}
 \hline
 Inequality & Label & Inequality & Label \\
 \hline
$w_1 \geqslant 0$ & \{1,1,1\} & $w_3 \leqslant  0.67 w_1$ & \{6,6,6\} \\
$w_2 \geqslant  0$ & \{2,2,2\} & $w_3 \geqslant  0.67 w_1$ & \{7,7,7\} \\
$w_3 \geqslant  0$ & \{3,3,3\} & $w_3 \geqslant  - 0.70 w_2 + 0.34 w_1 + 0.01$ & \{8,8,8\} \\
$w_3 \geqslant  - w_2 - w_1 + 1$ & \{4,4,4\} & $w_2 \leqslant  0.70 w_1 - 0.02$ & \{9,9,9\} \\
$w_3 \leqslant  - w_2 - w_1 + 1$ & \{5,5,5\} & $w_3 \geqslant  0.71 w_2 + 1.48 w_1 + 0.02$ & \{10,10,10\} \\
\hline
$w_2 \leqslant  - w_1 + 1$ & \{11,3,5\} & $w_2 \leqslant  - 4.49 w_1 + 3.31$ & \{15,8,5\} \\
$w_1 \geqslant  0$ & \{12,3,6\} & $w_2 \geqslant  - 0.46 w_1 + 0.02$ & \{16,8,6\} \\
$w_2 \geqslant  - 1.67 w_1 + 1$ & \{13,4,6\} & $w_2 \leqslant  - 1.45 w_1 + 0.57$ & \{17,10,5\} \\
$w_2 \leqslant  - 1.67 w_1 + 1$ & \{14,7,5\} & $w_2 \leqslant  - 1.15 w_1 - 0.02$ & \{18,10,6\} \\
\hline
$w_1 \geqslant  0.03$ & \{19,2,9\} & $w_1 \geqslant  1.97$ & \{28,13,17\} \\
$w_1 \leqslant  1.0$ & \{20,2,11\} & $w_1 \geqslant  1.97$ & \{29,13,18\} \\
$w_1 \leqslant  0.6$ & \{21,2,14\} & $w_1 \geqslant  0.04$ & \{30,16,9\} \\
$w_1 \leqslant  0.74$ & \{22,2,15\} & $w_1 \leqslant  1.83$ & \{31,16,11\} \\
$w_1 \leqslant  0.4$ & \{23,2,17\} & $w_1 \leqslant  0.82$ & \{32,16,14\} \\
$w_1 \leqslant  - 0.02$ & \{24,2,18\} & $w_1 \leqslant  0.82$ & \{33,16,15\} \\
$w_1 \geqslant  0.43$ & \{25,13,9\} & $w_1 \leqslant  0.56$ & \{34,16,17\} \\
$w_1 \geqslant  0.0$ & \{26,13,11\} & $w_1 \leqslant  - 0.06$ & \{35,16,18\} \\
$w_1 \leqslant  0.82$ & \{27,13,15\} & & \\ \hline
\end{tabular}
\end{table}

When verifying the truth of the~necessary preference relation for a~particular pair of alternatives, we consider an extended system of inequalities when compared with the one used for verifying the consistency between the DM's preference information and an assumed model (see Section~\ref{sec:uta_example_feasible}). In this perspective, the computational effort can be reduced by benefiting from the theoretical properties of the SD algorithm. Specifically, we can conduct solely operations involving the inequality related to the hypothesis on a~robust advantage of one alternative over another (\{10,10,10\}). When referring to the system of inequalities presented in Table~\ref{uta_necessary_true}, this would result only in the generation of inequalities \{17,10,5\} and \{18,10,6\}, and their subsequent crossover with \{2,2,2\}, \{13,13,13\}, and \{16,16,16\}. However, all other inequalities would remain the same, originating from the basic system of inequalities that was already transformed when checking whether compatible preference model instance ever exist.

\begin{figure}[!ht]
    \centering
    \includegraphics[scale=0.5]{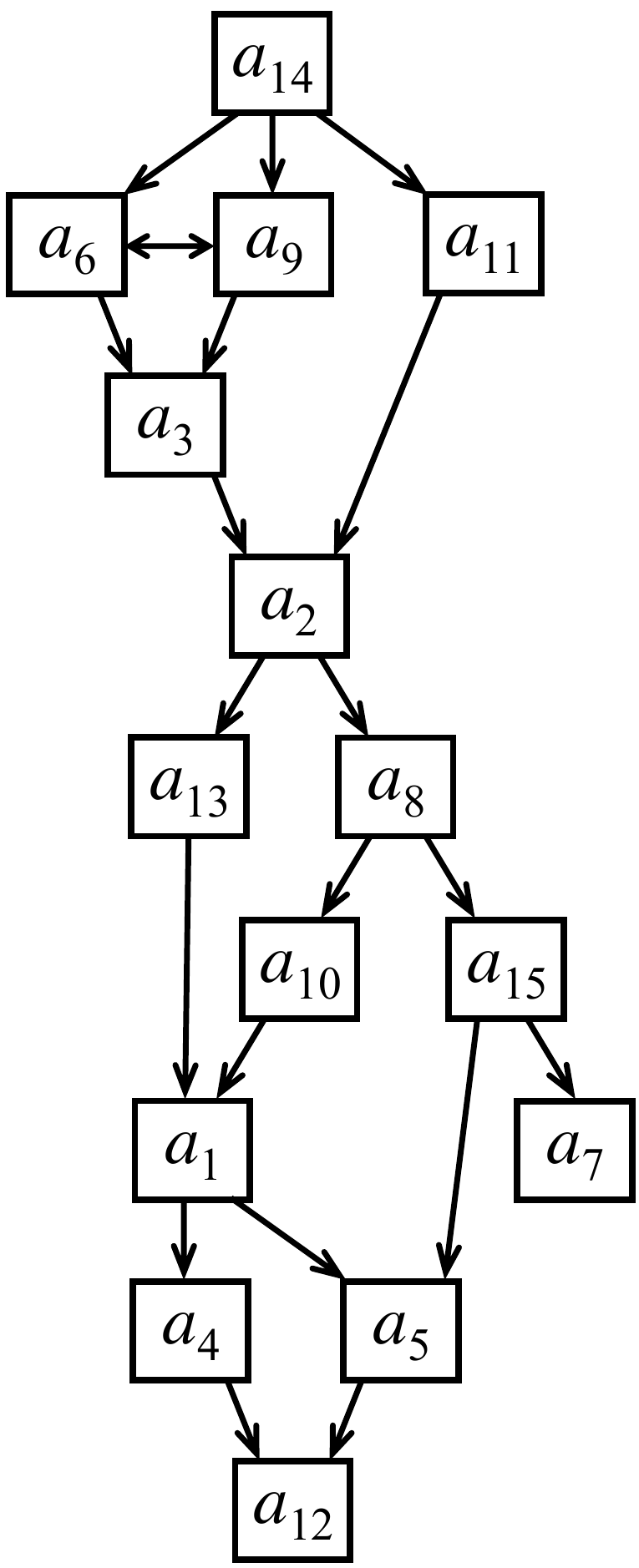}
    \caption{A Hasse diagram of the necessary preference relation (the arcs which can be obtained from the transitivity of the relation are omitted).}
    \label{fig:partia_preorder_necessary}
\end{figure}

Following the same way of reasoning, we verified the truth of the necessary preference relation for all pairs of alternatives. The resulting partial pre-order in form of a~Hasse diagram is presented in Figure~\ref{fig:partia_preorder_necessary}. The necessary relation is rich with $a_{14}$ being univocally preferred over all remaining alternatives, and $a_{12}$ being judged worse than other alternatives for all compatible preference model instances.

\subsubsection{Possible preference relation}
\label{sec:possible_weak}
\noindent The possible preference relation holds for a given pair of alternatives if one of them is not worst than the other for at least one compatible preference model instance~\citep{greco2008ordinal}. The first step of using the SD algorithm for verifying whether relation $a_i \succsim^P a_k$ can be instantiated consists of adding an inequality $U(a_i) \geqslant U(a_k)$ to the initial non-contradictory system modelling the DM's preferences. If~the resulting system does not involve any contradictions, then $U(a_i) \geqslant U(a_k)$ holds for at least one compatible preference model instance, and hence $a_i \succsim^P a_k$. Otherwise, $a_k$ is strictly preferred to $a_i$ for all compatible preference model instances, and thus $not(a_i \succsim^P a_k)$.

Let us consider alternatives $a_1$ and $a_2$. To check the truth of relation $a_1 \succsim^P a_2$, we need to extend an initial system from Table~\ref{example_init_uta_linear} with constraint $U(a_{1}) \geqslant U(a_{2})$ corresponding to the following inequality $w_3 \geqslant 0.1 w_2 + 1.11 w_1$ (\{10,10,10\}). The system obtained after the ISO and CRO operations is presented in Table~\ref{uta_possible_false}. It involves a few contradictions (e.g., \{28,13,17\} does not align with \{1,1,1\}, whereas \{24,2,18\} contradicts \{30,16,9\}). As a result, relation $a_1 \succsim^P a_2$ does not hold.

\begin{table}[!ht]
\centering\footnotesize
\caption{The SD system for verifying the truth of possible preference relation for pair $a_{1}$ and $a_{2}$ ($a_{1} \succsim^P a_{2}$).}
\label{uta_possible_false}
\begin{tabular}{lclc}
 \hline
 Inequality & Label & Inequality & Label \\
 \hline
$w_1 \geqslant 0$ & \{1,1,1\} & $w_3 \leqslant  0.67 w_1$ & \{6,6,6\} \\
$w_2 \geqslant 0$ & \{2,2,2\} & $w_3 \geqslant  0.67 w_1$ & \{7,7,7\} \\
$w_3 \geqslant 0$ & \{3,3,3\} & $w_3 \geqslant - 0.70 w_2 + 0.34 w_1 + 0.01$ & \{8,8,8\} \\
$w_3 \geqslant -  w_2 -  w_1 + 1$ & \{4,4,4\} & $w_2 \leqslant 0.70 w_1 - 0.02$ & \{9,9,9\} \\
$w_3 \leqslant - w_2 - w_1 + 1$ & \{5,5,5\} & $w_3 \geqslant 0.1 w_2 + 1.11 w_1$ & \{10,10,10\} \\
\hline
$w_2 \leqslant - w_1 + 1$ & \{11,3,5\} &$w_2 \leqslant - 4.49 w_1 + 3.31$ & \{15,8,5\}  \\
$w_1 \geqslant 0$ & \{12,3,6\} &  $w_2 \geqslant - 0.46 w_1 + 0.02$ & \{16,8,6\} \\
$w_2 \geqslant - 1.67 w_1 + 1$ & \{13,4,6\} & $w_2 \leqslant - 1.92 w_1 + 0.91$ & \{17,10,5\} \\
$w_2 \leqslant - 1.67 w_1 + 1$ & \{14,7,5\} & $w_2 \leqslant - 4.62 w_1$ & \{18,10,6\} \\
\hline
$w_1 \geqslant  0.03$ & \{19,2,9\}&$w_1 \leqslant - 0.34$ & \{28,13,17\}\\
$w_1 \leqslant  1$ & \{20,2,11\}&$w_1 \leqslant - 0.34$ & \{29,13,18\}\\
$w_1 \leqslant  0.60$ & \{21,2,14\}&$w_1 \geqslant  0.04$ & \{30,16,9\}\\
$w_1 \leqslant  0.74$ & \{22,2,15\} &$w_1 \leqslant  1.83$ & \{31,16,11\}\\
$w_1 \leqslant  0.47$ & \{23,2,17\}  &$w_1 \leqslant  0.82$ & \{32,16,14\}\\
$w_1 \leqslant  0$ & \{24,2,18\} &$w_1 \leqslant  0.82$ & \{33,16,15\}\\
$w_1 \geqslant  0.43$ & \{25,13,9\}  &$w_1 \leqslant  0.61$ & \{34,16,17\} \\
$w_1 \geqslant  0$ & \{26,13,11\} &$w_1 \leqslant 0$ & \{35,16,18\}\\
$w_1 \leqslant  0.82$ & \{27,13,15\}\\
\hline
\end{tabular}
\end{table}

When it comes to verification of $a_{2} \succsim^P a_{1}$, the process starts with extending an initial system (see Table~\ref{example_init_uta_linear}) with constraint $U(a_{2}) \geqslant U(a_{1})$ corresponding to the following inequality $w_3 \leqslant  0.1 w_2 + 1.11 w_1$ (\{10,10,10\}). The SD system obtained after processing the entire system of inequalities is presented in Table~\ref{uta_possible_true}. Since it is non-contradictory, $U(a_{2}) \geqslant U(a_{1})$ holds for at least one compatible preference model instance, which implies $a_{2} \succsim^P a_{1}$.

\begin{table}[!ht]
\centering\footnotesize
\caption{The SD system for verifying the truth of possible preference relation for pair $a_{2}$ and $a_{1}$ ($a_{2} \succsim^P a_{1}$).}
\label{uta_possible_true}
\begin{tabular}{lclc}
 \hline
 Inequality & Label & Inequality & Label \\
 \hline
    $w_3 \geqslant 0$ & \{3,3,3\} & $w_3 \leqslant - w_2 - w_1 + 1$ & \{5,5,5\} \\
    $w_3 \geqslant - w_2 - w_1 + 1$ & \{4,4,4\} & $w_3 \leq  0.67 w_1$ & \{6,6,6\} \\
    $w_3 \geqslant  0.67 w_1$ & \{7,7,7\} & $w_3 \leqslant  0.1 w_2 + 1.11 w_1$ & \{10,10,10\} \\
    $w_3 \geqslant - 0.7 w_2 + 0.34 w_1 + 0.01$ & \{8,8,8\} & \\
    \hline
    $w_2 \geqslant 0$ & \{2,2,2\} & $w_2 \leqslant 0.7 w_1 - 0.02$ & \{9,9,9\} \\
    $w_2 \geqslant - 11.61 w_1$ & \{13,3,10\} & $w_2 \leqslant - w_1 + 1$ & \{11,3,5\} \\
    $w_2 \geqslant - 1.67 w_1 + 1$ & \{14,4,6\} & $w_2 \leqslant - 1.67 w_1 + 1$ & \{16,7,5\} \\
    $w_2 \geqslant - 1.92 w_1 + 0.91$ & \{15,4,10\} & $w_2 \leqslant - 4.49 w_1 + 3.31$ & \{18,8,5\} \\
    $w_2 \geqslant - 4.62 w_1$ & \{17,7,10\} & \\
    $w_2 \geqslant - 0.46 w_1 + 0.02$ & \{19,8,6\} & \\
    $w_2 \geqslant - 0.96 w_1 + 0.02$ & \{20,8,10\} & \\
    \hline
    $w_1 \geqslant 0$ & \{1,1,1\} & $w_1 \leqslant + 0.6$ & \{23,2,16\} \\
    $w_1 \geqslant + 0.43$ & \{29,14,9\} & & \\ \hline
\end{tabular}
\end{table}

Following the same procedure, we checked the truth of possible preference relation for all pairs of alternatives. Table~\ref{tab:possible_weak_preference} indicates whether an alternative from the row is possibly preferred to an alternative from the column. Symbol ``T'' is used to indicate the truth of such a~relation (see, e.g., $(a_1, a_2)$), whereas ``F'' denotes its falsity (see, e.g., $(a_2, a_1)$).

\begin{table}[!ht]
\centering\footnotesize
\caption{A matrix of the possible preference relation (T -- true; F -- false).}
\label{tab:possible_weak_preference}
\begin{tabular}{cccccccccccccccc}
\hline
  & $a_{1}$ & $a_{2}$ & $a_{3}$ & $a_{4}$ & $a_{5}$ & $a_{6}$ & $a_{7}$ & $a_{8}$ & $a_{9}$ & $a_{10}$ & $a_{11}$ & $a_{12}$ & $a_{13}$ & $a_{14}$ & $a_{15}$ \\
  \hline
$a_{1}$ &  T&   F&   F&  T&  T&   F&  T&  T&   F&   F&   F&  T&   F&   F&  T\\
$a_{2}$ &  T&   T&   F&  T&  T&   F&  T&  T&   F&  T&   F&  T&  T&   F&  T\\
$a_{3}$ &  T&  T&   T&  T&  T&   F&  T&  T&   F&  T&  T&  T&  T&   F&  T\\
$a_{4}$ &  T&   F&   F&   T&  T&   F&  T&  T&   F&   F&   F&  T&   F&   F&  T\\
$a_{5}$ &  F&   F&   F&  T&   T&   F&  T&   F&   F&   F&  F &  T&   F&   F&  F \\
$a_{6}$ &  T&  T&  T&  T&  T&   T&  T&  T&  T&  T&  T&  T&  T&   F&  T\\
$a_{7}$ &  T&   F&   F&  T&  T&   F&   T&   F&   F&  T&   F&  T&  T&   F&  F \\
$a_{8}$ &  T&   F&   F&  T&  T&   F&  T&   T&   F&  T&   F&  T&  T&   F&  T\\
$a_{9}$ &  T&  T&  T&  T&  T&  T&  T&  T&   T&  T&  T&  T&  T&   F&  T\\
$a_{10}$ &  T&  F &   F&  T&  T&  F &  T&  T&  F &  T &  F &  T&  T&   F&  T\\
$a_{11}$ &  T&  T&  T&  T&  T&  T&  T&  T&  T&  T&   T&  T&  T&   F&  T\\
$a_{12}$ &   F&  F &  F &   F&   F&  F &  T&   F&  F &  F &   F&   T&  F &  F &  F \\
$a_{13}$ &  T&  F &  F &  T&  T&  F &  T&  T&  F &  T&  F &  T&   T&   F&  T\\
$a_{14}$ &  T&  T&  T&  T&  T&  T&  T&  T&  T&  T&  T&  T&  T&   T&  T\\
$a_{15}$ &  T&   F&  F &  T&  T&   F&  T&  F &   F&  T&   F&  T&  T&   F& T \\ \hline
\end{tabular}
\end{table}

\subsection{Generation of explanations}
\noindent In this section, we demonstrate how the SD can be employed to generate explanations of the provided recommendation. Selecting a single preference model instance or conducting robustness analysis prevents the DM from seeing the exact relations between preference information and results. In view of preference disaggregation approaches, two types of explanations may be of interest to the DM~\citep{Kadzinski14ORSP}. On the one hand, for some already observable result it is relevant to show the minimal set of preference information pieces underlying such an outcome. On the other hand, for some non-observable result it would be interesting to know the maximal set of DM's judgments admitting that such an outcome is instantiated. For illustrative purpose, we will justify some results discussed in Sections~\ref{sec:necessary_weak} and~\ref{sec:possible_weak}.

\subsubsection{Preference reduct}\label{sec:reduct}
\noindent A preference reduct is a minimal set of preference information pieces implying some outcome observable for all compatible preference model instances. As far as the setting considered in this paper is concerned, a preference reduct for a pair $(a_i, a_k) \in A \times A$, such that $a_i \succsim^N a_k$, is a minimal set of pairwise comparisons provided by the DM inducing $a_i \succsim^N a_k$ in the set of compatible value functions~\citep{Kadzinski14ORSP}. The employment of the SD algorithm for generating a~preference reduct for the necessary relation instantiated for a~given pair of alternatives starts with considering the final SD system used for verifying the truth of this relation. Let us remind that such a system -- justifying $a_i \succsim^N a_k$ -- needs to involve some contradictions. Then, all subsets of constraints leading to the contradiction need to be identified, and the labelling procedure should be used to indicate the constraints from the initial system underlying the contradiction. All these subsets would involve constraint $U(a_k) > U(a_i)$ along with constraints related to the formulation of employed preference model and/or pairwise comparisons provided by the DM. One needs to identify among them these constraints that involve the minimal subset(s) of DM's pairwise comparisons. The latter ones can be then judged as the preference reducts for $a_i \succsim^N a_k$.

Let us consider the necessary preference relation $a_{14} \succsim^N a_{1}$, whose validity was confirmed in Section~\ref{sec:necessary_weak}. The final SD system for verifying the truth of $a_{14} \succsim^N a_{1}$ is presented in Table~\ref{uta_necessary_true}. In this system, \{1,1,1\} -- \{5,5,5\} are model constraints, and the relation between the inequalities and the DM's pairwise comparisons is as follows: \{6,6,6\} and \{7,7,7\} -- $a_{6} \sim a_{9}$,  \{8,8,8\} -- $a_9 \succ a_8$, and \{9,9,9\} -- $a_8 \succ a_7$. Moreover, \{10,10,10\} is linked to the hypothesis $U(a_{14}) < U(a_{1})$ that could not be verified in the set of all compatible value functions, hence implying $a_{14} \succsim^N a_{1}$. In what follows, we will use the short symbols $1-10$ to refer to the basic constraints rather than their full forms \{1,1,1\} -- \{10,10,10\}. For example, constraint $1$ refers to $w_1 \geqslant 0$, whereas constraint $8$ is linked to $a_9 \succ a_8$.

To identify all contradictions, we let the SD algorithm  perform all CRO operations for variable $w_1$ rather than stopping at the first pair of contradictory ones. The reported inconsistencies are presented in Table~\ref{sets_of_contradictions}. For example, contradiction between $w_1 \geqslant 0$ (1 -- \{1,1,1\}) and $w_1 \leqslant  - 0.06$ (35 -- \{35,16,18\}) was implied by constraints 1, 6, 8, and 10 from the initial set of inequalities. When analyzing such subsets of constraints, we will ignore the constraints implied directly by the model ($1-5$) or the verified hypothesis ($10$). Instead, we will focus solely on the constraints implied by the DM's pairwise comparisons as only these can be considered as components of preference reducts. In this way, we identify the following subsets of constraints which contradict constraint $10$: \{6\}, \{6, 7\}, \{6, 8\}, \{6, 9\}, and \{6, 8, 9\}. Clearly, subset \{6\} is minimal among them. Therefore, the preference reduct for the necessary relation $a_{14} \succsim^N a_{1}$ contains just a single pairwise comparison, i.e., $a_{6} \sim a_{9}$.

\begin{table}[!ht]
\centering\footnotesize
\caption{Constraints leading to all contradictions in the SD system used for verifying the truth of $a_{14} \succsim^N a_{1}$.}
\label{sets_of_contradictions}
\begin{tabular}{ccl ccl}
 \hline
\multicolumn{2}{c}{Bounds} & Initial inequalities  &  \multicolumn{2}{c}{Bounds} & Initial inequalities\\
 \hline
 1 & 24 & [1, 2, 6, 10] & 28 & 32 & [4, 5, 6, 7, 8, 10] \\
1 & 35 & [1, 6, 8, 10]& 28 & 33 & [4, 5, 6, 8, 10] \\
12 & 24 & [2, 3, 6, 10] & 28 & 34 & [4, 5, 6, 8, 10] \\
12 & 35 & [3, 6, 8, 10] & 28 & 35 & [4, 5, 6, 8, 10] \\
19 & 24 & [2, 6, 9, 10] & 29 & 20 & [2, 3, 4, 5, 6, 10] \\
19 & 35 & [2, 6, 8, 9, 10] & 29 & 21 & [2, 4, 5, 6, 7, 10] \\
25 & 23 & [2, 4, 5, 6, 9, 10] & 29 & 22 & [2, 4, 5, 6, 8, 10] \\
25 & 24 & [2, 4, 6, 9, 10] & 29 & 23 & [2, 4, 5, 6, 10] \\
25 & 35 & [4, 6, 8, 9, 10] & 29 & 24 & [2, 4, 6, 10] \\
26 & 24 & [2, 3, 4, 5, 6, 10] & 29 & 27 & [4, 5, 6, 8, 10] \\
26 & 35 & [3, 4, 5, 6, 8, 10]& 29 & 31 & [3, 4, 5, 6, 8, 10] \\
28 & 20 & [2, 3, 4, 5, 6, 10] & 29 & 32 & [4, 5, 6, 7, 8, 10] \\
28 & 21 & [2, 4, 5, 6, 7, 10] & 29 & 33 & [4, 5, 6, 8, 10] \\
28 & 22 & [2, 4, 5, 6, 8, 10]& 29 & 34 & [4, 5, 6, 8, 10] \\
28 & 23 & [2, 4, 5, 6, 10] & 29 & 35 & [4, 6, 8, 10] \\
28 & 24 & [2, 4, 5, 6, 10] & 30 & 24 & [2, 6, 8, 9, 10] \\
28 & 27 & [4, 5, 6, 8, 10] & 30 & 35 & [6, 8, 9, 10] \\
28 & 31 & [3, 4, 5, 6, 8, 10] \\
\hline
\end{tabular}
\end{table}

\subsubsection{Preference construct}
\label{sec:construct}
\noindent A preference construct is a maximal set of preference information pieces admitting the truth of some outcome non-observable for all compatible preference model instances. When referring to the setting considered in this paper, a preferential construct for a pair $(a_i, a_k) \in A \times A$, such that $not(a_i \succsim^P a_k)$, is a maximal set of pairwise comparisons provided by the DM admitting $a_i \succsim^P a_k$~\citep{Kadzinski14ORSP}. The employment of the SD algorithm for generating a preference construct for the possible relation to be instantiated for a given pair of alternatives starts with considering the final SD system used for verifying the truth of this relation. Let us remind that such a system needs to involve some contradictions so that the falsity of $a_i \succsim^P a_k$ is justified. Thus, similarly to the computation of a~preference reduct, we need identify all subsets of constraints implying the contradiction with $U(a_i) \geqslant U(a_k)$, link them to the initial constraints, and find only the minimal subsets referring of DM's pairwise comparisons, which do not align with $U(a_i) \geqslant U(a_k)$. The remaining pairwise comparisons are then deemed as the preference construct for $a_i \succsim^P a_k$.

Let us consider the possible preference relation $a_1 \succsim^P a_2$, whose falsity was proved in Section~\ref{sec:possible_weak}. The final SD system which leads to claiming that $not(a_{1} \succsim^P a_{2})$ is presented in Table~\ref{uta_possible_false}. The interpretation of constraints \{1,1,1\} -- \{9,9,9\} is the same as in Section~\ref{sec:reduct}, whereas \{10,10,10\} is linked to the hypothesis $U(a_{1}) \geqslant U(a_{2})$ that could not be verified in the set of all compatible value functions, thus implying $not(a_{1} \succsim^P a_{2})$. All subsets of constraints leading to the contradiction are reported in Table~\ref{sets_of_contradictions}. Having ignored constraints $1-5$, we identify the following subsets of constraints related to the DM's pairwise comparisons which contradict constraint $10$: \{6\},  \{6, 8\}, \{6, 9\}, and \{6, 8, 9\}. Clearly, subset \{6\} is minimal among them, and hence $a_1 \succsim^P a_2$ would be true once pairwise comparison $a_{6} \succsim a_{9}$ is eliminated. Thus, the preference construct for the possible preference relation $a_{1} \succsim^P a_{2}$ consists of the following pairwise comparisons: $a_9 \succ a_6 \succ a_8 \succ a_7$.

\begin{table}[!ht]
\centering\footnotesize
\caption{Constraints leading to all contradictions in the SD system used for verifying the truth of $a_{1} \succsim^P a_{2}$.}
\label{sets_of_contradictions_2}
\begin{tabular}{ccl ccl}
 \hline
\multicolumn{2}{c}{Bounds} & Initial inequalities  &  \multicolumn{2}{c}{Bounds} & Initial inequalities\\
 \hline
1 & 28 & [1, 4, 5, 6, 10] & 25 & 28 & [4, 5, 6, 9, 10] \\
1 & 29 & [1, 4, 6, 10] & 25 & 29 & [4, 6, 9, 10] \\
1 & 35 & [1, 6, 8, 10] & 25 & 35 & [4, 6, 8, 9, 10] \\
12 & 28 & [3, 4, 5, 6, 10] & 26 & 28 & [3, 4, 5, 6, 10] \\
12 & 29 & [3, 4, 6, 10] & 26 & 29 & [3, 4, 5, 6, 10] \\
12 & 35 & [3, 6, 8, 10] & 26 & 35 & [3, 4, 5, 6, 8, 10] \\
19 & 24 & [2, 6, 9, 10] & 30 & 24 & [2, 6, 8, 9, 10] \\
19 & 28 & [2, 4, 5, 6, 9, 10] & 30 & 28 & [4, 5, 6, 8, 9, 10] \\
19 & 29 & [2, 4, 6, 9, 10] & 30 & 29 & [4, 6, 8, 9, 10] \\
19 & 35 & [2, 6, 8, 9, 10] & 30 & 35 & [6, 8, 9, 10] \\
25 & 24 & [2, 4, 6, 9, 10] \\
\hline
\end{tabular}
\end{table}

\subsection{Generation of holistic preference criteria reducts}
\noindent In this section, we show how the SD algorithm can be used to determine the holistic preference criteria reducts. We define them as minimal subsets of criteria that -- when incorporated into the assumed preference model -- would reproduce the DM's holistic judgments. The verification procedure consists of a suitable adaptation of an additive method. That is, we model the SD system for each subset of criteria, starting with these with the least cardinality. Once a certain SD system is non-contradictory, a respective subset of criteria can be considered as a holistic preference criteria reduct, and its proper supersets are not verified, because they would not satisfy the requirement of non-redundancy. The procedure is continued until all subsets of criteria are considered.

Let us demonstrate the results of a verification procedure for a few selected subsets of criteria. When using solely $g_1$ to reproduce the DM's reference ranking, the initial SD system is presented in Table~\ref{min_sets_only_1}.  It involves several contradictions (e.g., \{1,1,1\} does not align with \{6,6,6\}, whereas \{2,2,2\} contradicts \{5,5,5\}). Therefore, $g_1$ cannot be considered as a holistic preference criteria reduct. Similarly, neither $g_2$ nor $g_3$ is sufficient for reproducing the DM's reference ranking.

\begin{table}[!ht]
\centering\footnotesize
\caption{Initial system for validation of criterion $g_1$ as a holistic preference criteria reduct.}
\label{min_sets_only_1}
\begin{tabular}{lclc}
 \hline
 Inequality & Label & Inequality & Label \\
 \hline
$w_1 \geqslant 0.0$ & \{1,1,1\} & $w_1 \leqslant 1.0$ & \{3,3,3\} \\
$w_1 \geqslant 1.0$ & \{2,2,2\} & $w_1 \leqslant 0.0$ & \{5,5,5\} \\
$w_1 \geqslant 0.0$ & \{4,4,4\} & $w_1 \leqslant - 0.04$ & \{6,6,6\} \\
$w_1 \geqslant 0.03$ & \{7,7,7\} & & \\
\hline
\end{tabular}
\end{table}

Let us now go to the verification whether some pairs of criteria ensure compatibility with the DM's holistic judgments. In Table~\ref{min_sets_only_23}, we present the set of inequalities obtained by the SD algorithm whose scope was limited only to criteria $g_2$ and $g_3$. It is contradictory (see, e.g., \{1,1,1\} and \{8,8,8\} or \{10,3,8\} and \{5,5,5\}), and hence $\{g_2, g_3\}$ is not a holistic preference criteria reduct. In turn, the SD system involving only criteria $g_1$ and $g_3$ is not contradictory (see Table~\ref{min_sets_only_13}), and $\{g_1, g_3\}$ is a minimal subset of criteria able to reproduce the DM's preference ranking. Indeed, it is the sole holistic preference criteria reduct for the assumed preference model and information.

\begin{table}[!ht]
\centering\footnotesize
\caption{The SD system for validation of a subset of criteria $\{g_2,g_3\}$ as a holistic preference criteria reduct.}
\label{min_sets_only_23}
\begin{tabular}{lclc}
 \hline
 Inequality & Label & Inequality & Label \\
 \hline
$w_2 \geqslant 0.0$ & \{1,1,1\} & $w_2 \leqslant -1.0 w_3 + 1.0$ & \{4,4,4\} \\
$w_2 \geqslant -1.0 w_3 + 1.0$ & \{3,3,3\} & $w_2 \leqslant -0.02$ & \{8,8,8\} \\
$w_2 \geqslant -1.43 w_3 + 0.02$ & \{7,7,7\} & & \\
\hline
$w_3 \geqslant 0.0$ & \{2,2,2\} & $w_3 \leqslant 0.0$ & \{5,5,5\} \\
$w_3 \geqslant 0.0$ & \{6,6,6\} & $w_3 \leqslant 1.0$ & \{9,1,4\} \\
$w_3 \geqslant 1.02$ & \{10,3,8\} & & \\
$w_3 \geqslant -2.31$ & \{11,7,4\} & & \\
$w_3 \geqslant 0.03$ & \{12,7,8\} & & \\
\hline
\end{tabular}
\end{table}

\begin{table}[!ht]
\centering\footnotesize
\caption{The SD system for validation of a subset of criteria $\{g_1,g_3\}$ as a holistic preference criteria reduct.}
\label{min_sets_only_13}
\begin{tabular}{lclc}
 \hline
 Inequality & Label & Inequality & Label \\
 \hline
$w_3 \geqslant 0.0$ & \{2,2,2\} & $w_3 \leqslant - 1.0 w_1 + 1.0$ & \{4,4,4\} \\
$w_3 \geqslant - 1.0 w_1 + 1.0$ & \{3,3,3\} & $w_3 \leqslant 0.67 w_1 + 0.0$ & \{5,5,5\} \\
$w_3 \geqslant 0.67 w_1 + 0.0$ & \{6,6,6\} & & \\
$w_3 \geqslant 0.34 w_1 + 0.01$ & \{7,7,7\} & & \\
 \hline
$w_1 \geqslant 0.0$ & \{1,1,1\} & $w_1 \leqslant 0.6$ & \{12,6,4\} \\
$w_1 \geqslant 0.03$ & \{8,8,8\} & $w_1 \leqslant 0.74$ & \{13,7,4\} \\
$w_1 \geqslant 0.0$ & \{10,2,5\} & $w_1 \leqslant 1.0$ & \{9,2,4\} \\
$w_1 \geqslant 0.6$ & \{11,3,5\} & & \\
$w_1 \geqslant 0.04$ & \{14,7,5\} & & \\
\hline
\end{tabular}
\end{table}


\section{Concluding remarks}
\label{sec:conslusions_remarks}
\noindent In this paper, we proposed a new method for analyzing a set of parameters used in a multiple criteria ranking method. Unlike the existing techniques, we did not use any optimization technique, instead incorporating and extending a SD approach. While considering a value-based preference disaggregation method, we validated the proposed algorithm in a~multi-purpose analysis including inconsistency treatment, robustness preoccupation, and generation of explanations. Its practical usefulness was exemplified on a numerical study.

First, we revealed the suitability of SD approach for verifying the consistency between the DM's preference information and an assumed preference model. In this perspective, we showed that it could be used for identifying the sources of potential inconsistency by capturing a~set of constraints associated with the contradiction in the system of inequalities. Such information can be analyzed by the Decision Maker to revise his or her judgment policy. This refers to a~more general possibility of incorporating the method for verification if a problem for which the space of potential solutions is modelled with mathematical constraints can be ever solved.

Second, we illustrated how the SD algorithm can be used for robustness analysis in case a set of parameters compatible with the provided preference information is non-empty. In this perspective, the method can be used to discover the extreme compatible values of parameters (e.g., trade-off weights related to the formulation of marginal value functions). Moreover, due to constructing a system of interrelated inequalities, the algorithm may be used to emphasize an impact of some variable on the compatible values of other parameters (e.g., an admissible weight of one criterion vs. the weights of the remaining criteria). In addition, we demonstrated that it is suitable for discovering all compatible sets of parameters which can be subsequently exploited in a convenient way. Specifically, we showed how the method can be employed for verifying the validity of necessary and possible results confirmed by, respectively, all or at least one compatible sets of parameters. In this context, the usefulness of the proposed approach is enhanced by the internal nature of robustness analysis, which consists of the verification of various hypotheses within the same system of inequalities. In~fact, once the basic system is processed, the extended one can be easily handled without repeating the operations for the constraints involved in the initial system.

Third, we made evident how the analysis of inconsistency with the SD approach can be used for generating arguments about the validity of results and the role of particular criteria. In this perspective, we focused on computing the preference reducts and constructs, which denote, respectively, the minimal subset of preference information pieces implying the truth of some outcome or the maximal subset of such pieces admitting the validity of some currently non-observable result. Moreover, we introduced the concept of a~holistic preference criteria reduct corresponding to a~minimal subset of criteria sufficient for reproducing the DM's preferences.

The major challenge related to the use of SD concerns a~large number of crossover operations to be conducted when processing all variables. In the context of UTA, it depends on the numbers of criteria, characteristic points, and pairwise comparisons. However, this problem can be partially addressed by eliminating the redundant constraints when processing each variable. Moreover, the scenarios considered in MCDA usually involve only up to a~few criteria, the assumed preference models are assumed to be as simple as possible for increasing their interpretability~\citep{KADZINSKI2017146}, whereas the collections of pairwise comparisons provided by the Decision Maker are modestly sized~\citep{KadzinskiML}.

We envisage the following directions for future research. As far as inconsistency analysis is concerned, so far we focused only on studying the role of holistic judgments, while neglecting the constraints related to the assumptions of a~preference model. Thus, instead of revising the preference information, the DM and a decision analyst may be advised to change or adjust the model so that to increase its expressiveness~\citep{KADZINSKI2017146}. When considering an~additive value function this can be attained, e.g., by means of admitting non-monotonic marginal value functions~\citep{GHADERI20171073} or inverting a~preference direction for a~given criterion.

Furthermore, since the Segmenting Description algorithm can be used to analyze any set of inequalities, it can be employed in the context of other decision aiding methods. The most appealing extensions concern the preference disaggregation methods for multiple criteria sorting methods such as UTADIS~\citep{Devaud80, Zopounidis00} ELECTRE TRI~\citep{DIAS2002332, Kadzinski15, SILVA2016275}, and FlowSort~\citep{NEMERY201554, CAMPOS2015115}. Also, a~promising feature of SD, making it superior to LP, derives from its capability of handling non-linear constraints. The latter emphasizes the method's suitability for handling non-standard preferences~\citep{MARIN20135} and more complex models.

\section*{Acknowledgements}
\addcontentsline{toc}{section}{\numberline{}Acknowledgements}
\noindent Mi\l{}osz Kadzi\'{n}ski acknowledges the support from the Polish Ministry of Science and Higher Education (grant number 0296/IP2/2016/74). Jos\'{e} Rui Figueira acknowledges the support from the Funda\c c\~ao para a~Ci\^encia e Tecnologia (FCT) (grant number SFRH/BSAB/139892/2018) under POCH Program and the FCT for supporting the WISDom project (reference number DSAIPA/DS/0089/2018), and for partially funding the current research.

\vfill\newpage


\addcontentsline{toc}{section}{\numberline{}References}
\bibliographystyle{model2-names}
\bibliography{references}

\end{document}